\documentclass[letterpaper, 10 pt, conference]{class/ieeeconf}  

\IEEEoverridecommandlockouts                              
\usepackage{siunitx}
\usepackage[table,x11names]{xcolor}
\usepackage{graphicx}

\usepackage{amsmath}
\usepackage{pifont}
\usepackage{tcolorbox}
\usepackage{enumerate}
\let\labelindent\relax
\usepackage[shortlabels]{enumitem}\usepackage{amssymb}
\usepackage{latexsym}
\usepackage{url}
\usepackage{cite}
\usepackage{relsize}
\usepackage{multirow}
\usepackage{afterpage}
\usepackage{ifthen}
\usepackage{graphicx}
\usepackage{algpseudocode,algorithm,algorithmicx}
\usepackage{subfigure}
\usepackage{flushend}
\usepackage{epstopdf}
\usepackage{stackengine}
\usepackage{textcomp} 
\usepackage{pifont}
\usepackage{threeparttable,booktabs}
\usepackage{gensymb}
\usepackage{booktabs}
\usepackage{makecell}
\usepackage{hhline}
\usepackage{courier}
\usepackage{lipsum}
\usepackage{diagbox}
\usepackage{xcolor} 
\usepackage{xspace}
\usepackage{bbm}

\usepackage{xspace}

\newcommand{\nbf}[1]{{\noindent \textbf{#1}}}

\makeatletter
\let\NAT@parse\undefined
\makeatother
\usepackage[bookmarks=false, linkcolor=blue, urlcolor=blue, citecolor=blue]{hyperref} 
 \graphicspath{{./figures/}}

\hypersetup{
    colorlinks=true,
    linkcolor=red,
    filecolor=magenta,      
    urlcolor=blue,
    pdfstartview={FitH},
    citecolor =blue
    }

\title{\LARGE \bf
Action Tokenizer Matters in In-Context Imitation Learning
}

\author{An Dinh Vuong$^{1}$, Minh Nhat Vu$^{2}$, Dong An$^{1}$, Ian Reid$^{1}$
\thanks{$^{1}$Department of Computer Vision,
        Mohammed bin Zayed University of Artificial Intelligence, UAE
        {\tt\small an.vuong@mbzuai.ac.ae}
        }%
\thanks{$^2$ AIT Austrian Institute of Technology GmbH, Austria  
}
}

\begin{document}

\maketitle
\thispagestyle{empty}
\pagestyle{empty}

\begin{abstract}
In-context imitation learning (ICIL) is a new paradigm that enables robots to generalize from demonstrations to unseen tasks without retraining. 
A well-structured action representation is the key to capturing demonstration information effectively, yet action tokenizer (the process of discretizing and encoding actions) remains largely unexplored in ICIL.
In this work, we first systematically evaluate existing action tokenizer methods in ICIL and reveal a critical limitation: while they effectively encode action trajectories, they fail to preserve temporal smoothness, which is crucial for stable robotic execution. 
To address this, we propose LipVQ-VAE, a variational autoencoder that enforces the Lipschitz condition in the latent action space via weight normalization. By propagating smoothness constraints from raw action inputs to a quantized latent codebook, LipVQ-VAE generates smoother actions. 
When integrating into ICIL, LipVQ-VAE improves performance by more than 5.3\% in high-fidelity simulators, with real-world experiments confirming its ability to produce smoother, more reliable trajectories. Code and checkpoints are available at \href{https://action-tokenizer-matters.github.io/}{https://action-tokenizer-matters.github.io/}.

\end{abstract}


\section{INTRODUCTION}
Imitation Learning (IL) is a popular training paradigm in robotic manipulation, where models learn from collected demonstrations to generalize across tasks~\cite{mandlekar2021matters}. 
A fundamental challenge of IL is limited generalization~\cite{team2024octo}, primarily due to the scarcity of high-quality demonstrations. 
While large-scale datasets, such as Octo~\cite{team2024octo}, Open X-Embodiment~\cite{o2024open}, and DROID~\cite{khazatsky2024droid}, have been introduced to mitigate this issue, adapting to novel tasks still requires collecting additional task-specific demonstrations for fine-tuning, restricting practical applications.
To address this, in-context imitation learning (ICIL) has recently emerged~\cite{di2024keypoint, xia2024kinematic, vosylius2025instant}.
Inspired by the attractive in-context learning capabilities of large language models (LLMs)~\cite{wang2024large}, ICIL enables robots to infer and execute new tasks directly from expert demonstrations without retraining~\cite{fu2024context}. 
ICIL holds significant potential for enhancing generalization in robotic learning, paving the way for more flexible and efficient real-world deployment.

\begin{figure}[!ht]
    \centering
    \includegraphics[width=\linewidth]{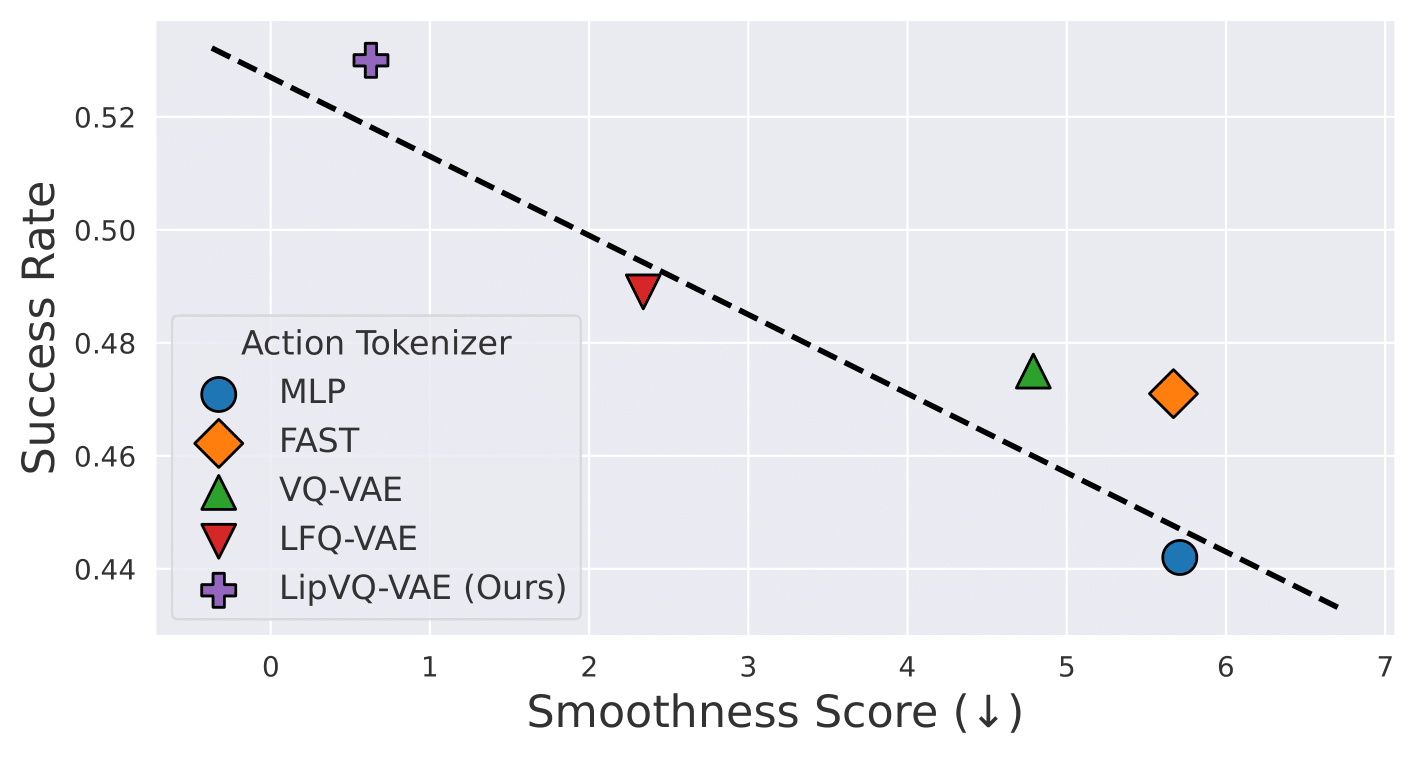}
    \caption{We examine the impact of action tokenizer on in-context imitation learning. Our findings indicate that a smoother representation of action correlates with higher robotic manipulation success. \textbf{Note in the figure:} a lower smoothness score reflects a smoother action representation.}
    \label{fig:teaser}
\end{figure}
Despite its potential, current ICIL research hardly focuses on learning contextual representation from rich demonstrations~\cite{park2025incontext}. 
In particular, most works overlook \textit{action representation}, a key component of contextual demonstrations, even though an effective contextual representation mechanism can significantly improve in-context learning performance~\cite{wang2024large}. Most works rely on LLMs, which struggle with precise, long-horizon motion planning due to inaccurate action representation and latency~\cite{wang2025skil}. To bridge this gap, we study the role of action representation in ICIL for robotic manipulation, aiming to improve in-context learning through effective action tokenization.

Discretizing complex, continuous action spaces is fundamental to sequential decision-making in robotic manipulation~\cite{pertsch2025fast, zheng2024prise}. 
Consequently, robust action tokenization remains an active research area~\cite{zhao2023learning, song2023lipsnet, brohan2022rt, wang2024smooth, yang2024vq, fujimoto2024sale}, with a key challenge being the handling of temporal correlations~\cite{pertsch2025fast}. While preserving the temporal order of an action trajectory can be effectively solved with positional encoding in transformers~\cite{vaswani2017attention}, or vector quantization~\cite{yu2024language}; a critical gap in ICIL remains: ensuring \textit{temporal smoothness}~\cite{bharadhwaj2024roboagent}. The temporal smoothness maintains continuity between manipulation outputs, thereby reducing noise and modeling uncertainties~\cite{mysore2021regularizing}. We empirically find a correlation between smooth action representation and successful robotic manipulation. As summarized in Fig.~\ref{fig:teaser}, improved smoothness in action representation leads to enhanced manipulation performance. 

This paper overcomes a largely overlooked challenge in learning contextual representation within ICIL  by introducing LipVQ-VAE, an action tokenizer that utilizes Lipschitz regularization through a simple yet effective weight normalization approach~\cite{liu2022learning}. Built upon vector quantization~\cite{van2017neural}, a powerful unsupervised representation learning technique, LipVQ-VAE can be integrated with contextual tokens from autoregressive models, such as transformers~\cite{vaswani2017attention}, to enable in-context learning~\cite{ma2024survey} effectively. 
We empirically show that conventional action tokenizers in imitation learning~\cite{zhao2023learning, pertsch2025fast} fail to preserve temporal smoothness. In contrast, LipVQ-VAE improves temporal smoothness by enforcing the Lipschitz condition into the latent space~\cite{duan2022temporal}, improving landscape smoothness and noise robustness~\cite{song2023lipsnet}. 
Our contributions are three-fold:
\begin{itemize}[leftmargin=*]
    \item To our knowledge, this work makes the first exploration of the role of action tokenization in ICIL, with empirical results highlighting its significance on ICIL’s success.
    \item We identify temporal smoothness as a key limitation of existing action tokenizers in ICIL and introduce LipVQ-VAE, a quantization-based tokenizer that enforces Lipschitz constraints to address this limitation.
    \item Experiments on RoboCasa~\cite{nasiriany2024robocasa}, ManiSkill~\cite{tao2024maniskill3}, and real robots indicate that LipVQ-VAE surpasses other action tokenizers, facilitating more robust ICIL performance.
\end{itemize}
\section{RELATED WORK}
\label{sec:related_work}
\nbf{In-context Imitation Learning for Robotic Manipulation.} 
ICIL has recently attracted significant interest in robotic manipulation in both simulators and real robots~\cite{wang2025skil}, demonstrating efficiency in enhancing policy generalization by effectively utilizing training demonstrations~\cite{fu2024context}. 
Pioneering ICIL methods exploit the inherent in-context learning (ICL) capabilities of large language models (LLMs)~\cite{brown2020language,mirchandani2023large,kwon2024language, vosylius2025instant,vosylius2023few}.
These approaches explicitly encode scene and robot states as keypoint-based representations~\cite{di2024keypoint}, allowing LLMs to infer action based on contextual information in the form of state dependencies.
However, these methods often rely on external vision foundation models, such as keypoint detection~\cite{caron2021emerging}, which lacks inherent generalizability across diverse tasks and object interactions~\cite{xu2024flow}. 

Recently, researchers have begun integrating ICIL into a trainable framework, leveraging causal transformers to model ICIL as a next-token prediction problem~\cite{fu2024context,zhang2025dynamics}. In this paradigm, both scene and action representations are jointly optimized, enabling more flexible and scalable policy learning. Building on this line of research, our work systematically investigates the role of action tokenization in ICIL and introduces a smoothness-aware action tokenizer to improve ICIL's performance.

\nbf{Learnable Action Tokenizer.} 
There are numerous ways to represent actions in robotics~\cite{chandak2019learning}, and capturing action is a long-standing substantial problem for robotics research~\cite{zech2019action}. 
From a machine learning perspective, action representation can be viewed as a feature learning~\cite{fujimoto2024sale}. Broadly, there are four learning approaches: \textit{(i)} discretization, where each dimension of the action is divided into discrete bins~\cite{shafiullah2022behavior, brohan2022rt}; \textit{(ii)} function approximation, where actions are parameterized by a neural network~\cite{fu2024context, zhang2025dynamics}; \textit{(iii)} latent space representation, where actions are encoded in a lower-dimensional space using unsupervised learning techniques such as variational autoencoders (VAE)~\cite{zhao2023learning}; and \textit{(iv)} sequence of language tokens~\cite{pertsch2025fast, kim2024openvla, huang2024emotion}, widely used in vision-language-action models~\cite{o2024open}. 
While discretization and neural approximation effectively capture the action representation~\cite{alwani2022decore}, latent space representations often exhibit posterior training generalization~\cite{allshire2021laser}, making them well-suited for ICIL. 
Language-based action tokenizer shows promise for vision-and-language foundation models~\cite{team2024octo}, though integrating a joint contextual representation across multiple modalities remains a key challenge~\cite{ma2024survey}.

\nbf{Smooth Action Regularization.}  
Smooth action representation is crucial for stabilizing control signals in robotic manipulation~\cite{ mysore2021regularizing}, promoting extensive research in the field. 
Watson \textit{et al.} employ Gaussian process priors to infer posterior actions, ensuring the trajectory smoothness while preserving performance and sample efficiency~\cite{watson2023inferring}. CAPS enforces smoothness by approximating a Lipschitz condition in the policy network~\cite{mysore2021regularizing}. 
ACT introduces temporal ensembling by aggregating action over a fixed \textit{chunk} size~\cite{zhao2023learning}. 
However, ACT still relies on a VAE-based tokenizer, and the problem of preserving temporal smoothness in the latent space is not explicitly investigated in~\cite{zhao2023learning}. 
Styrud \textit{et al.} formulate manipulation as Behavior Trees, leveraging Bayesian optimization for action smoothing~\cite{styrud2024bebop}. 
This work revisits Lipschitz regularization, motivated by its simplicity and effectiveness in enforcing latent space smoothness~\cite{liu2022learning}.

\section{METHOD}
This section introduces a smoothness-enhanced action tokenizer for ICIL.
We first outline the preliminaries of ICIL based on the next-token prediction framework~\cite{fu2024context}.
Then, we describe the proposed LipVQ-VAE tokenizer, which improves action representation smoothness by incorporating the Lipschitz condition into the variational encoder network~\cite{song2023lipsnet}.

\subsection{Preliminaries on In-Context Imitation Learning}
\label{subsec: ICRT}

Unlike traditional imitation learning, in-context imitation learning (ICIL) leverages the current scene observation and additional contextual information, such as \textit{prompt demonstrations.} 
We adopt the ICRT framework~\cite{fu2024context}, forming robotic manipulation as a next-token prediction problem. 
ICRT integrates contextual information by encoding a batch of prompt demonstrations, jointly representing observations and actions within a shared latent token space. These joint observation-action tokens serve as a prompt, guiding the policy to predict actions for new observations. 
This paper aims to investigate action encoding within ICRT, emphasizing that the choice of action tokenizer—which encodes actions at each timestep—plays a crucial role in ICIL performance.

\begin{figure}[!ht]
    \centering
    \includegraphics[width=\linewidth,trim=47 5 47 0,clip]{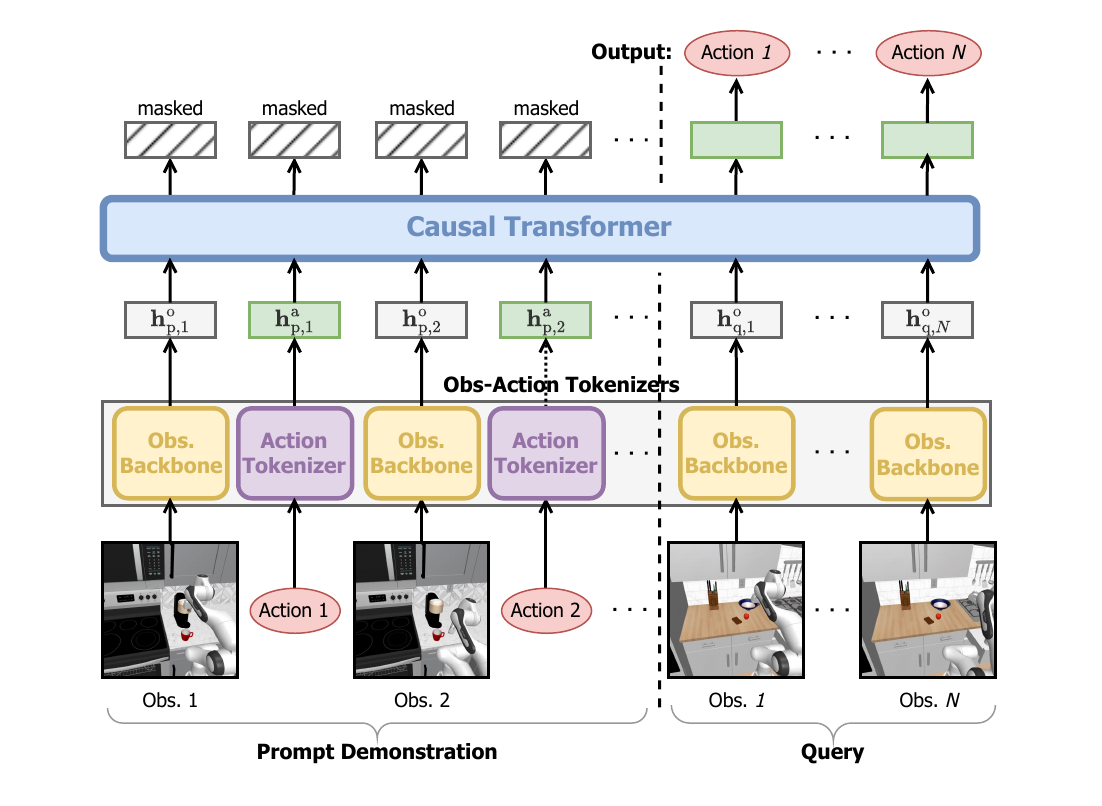}
    \caption{\textbf{ICRT architecture.} ICRT models action prediction as the next-token generation, utilizing prompt demonstrations for in-context learning.}
    \label{fig:icrt-architecture}
\end{figure}

The overview of the ICRT framework is shown in Fig.~\ref{fig:icrt-architecture}. A key distinction from conventional imitation learning approaches is the division of expert demonstrations into two components: the prompt demonstrations and the query. Observations and actions within the prompt demonstrations are tokenized into prompt tokens, providing the model with contextual information on executing the robotic task. An autoregressive transformer subsequently processes these tokens to generate action predictions for a new sequence of query observations. The model can generalize to unseen tasks without retraining by conditioning on prompt demonstrations.

All observations and actions are first processed through a layer of tokenizers consisting of two components: an observation backbone and an action tokenizer. The observation backbone employs encoder networks to process sensory inputs and language conditions. Following standard policy implementations, RGB images from RoboCasa~\cite{nasiriany2024robocasa} and RGB-D images from ManiSkill~\cite{tao2024maniskill3} are encoded using ResNet-18~\cite{he2016deep}, while language condition is embedded via CLIP~\cite{radford2021learning}. Other sensory data from the simulators are processed with MLPs. 
Tokenized observations and actions are then projected into a shared latent space, where another MLP ensures the same dimension between the two components.


Causal transformers are one of the most suitable architectures for ICIL due to their autoregressive nature, ensuring that each token depends only on previous tokens~\cite{brown2020language}. 
The transformer processes a sequence of observation tokens $\mathbf{h}_*^\text{o}$ and action tokens $\mathbf{h}_*^\text{a}$ generated in the previous step:
\begin{equation}
    \label{eq: icrt-sequence}\underbrace{(\mathbf{h}_{\text{p},1}^\text{o}, \mathbf{h}_{\text{p},1}^\text{a}, \dots, \mathbf{h}_{\text{p},M}^\text{o}, \mathbf{h}_{\text{p},M}^\text{a})}_{\text{prompt}},
\underbrace{(\mathbf{h}_{\text{q},1}^\text{o}, \mathbf{h}_{\text{q},2}^\text{o}, \dots, \mathbf{h}_{\text{q},N}^\text{o})}_{\text{query}},
\end{equation}
where $M$ is the number of timesteps in the prompt demonstrations, and $N$ is the number of timesteps in the query. Following~\cite{fu2024context}, we use one full demonstration of a robotic manipulation task as the prompt. The transformer outputs tokens only for the query while masking those corresponding to the prompt. Non-masked tokens are decoded via MLPs to generate actions for the query observations. The network is trained by supervising predicted actions with ground truth actions, following the behavior cloning approach in~\cite{nasiriany2024robocasa}. During inference, the model processes only one query observation at a time and predicts the action autoregressively.


In the original ICRT framework~\cite{fu2024context}, they use a simple MLP as the action tokenizer.
However, the na\"ive MLP encoding may lack sufficient smoothness in the action latent space. 
Smoothness is crucial for the success of ICIL, as our analysis reveals a correlation between smoother action representation and higher manipulation success rates (see Fig.~\ref{fig:teaser} and more details are in Sec.~\ref{sec: experiment}).
We propose a smoothness-enhanced action tokenizer to address this limitation, detailed in Sec.~\ref{sec:lipvq}.

\subsection{Lipschitz-Constrained Smooth Action Tokenizer}\label{sec:lipvq}
\begin{figure}[!ht]
    \centering
    \includegraphics[width=0.95\linewidth, trim=0 0 110 0, clip]{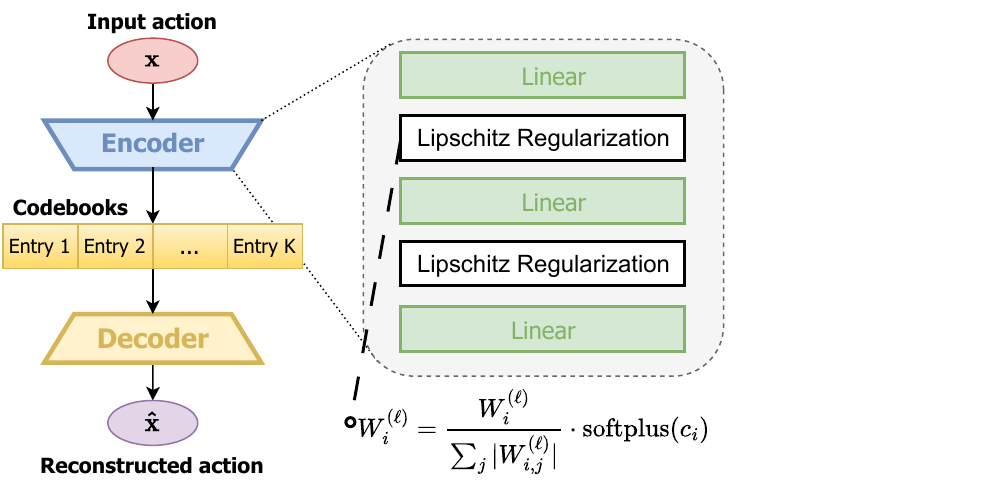}
    \caption{\textbf{LipVQ-VAE action tokenizer overview.} We adopt an autoencoder framework that maps actions to a latent space via codebook lookup. To ensure smooth latent representation, we apply Lipschitz regularization by row-wise normalizing the weight matrix after each encoder layer.}
    \label{fig:lipfq-vae architecture}
\end{figure}

For better action tokenization, we introduce \textbf{Lip}schitz-Constrained \textbf{V}ector \textbf{Q}uantization \textbf{V}ariational \textbf{A}uto\textbf{E}ncoder (LipVQ-VAE).
LipVQ-VAE is an autoencoder that encodes actions into a latent space using vector quantization~\cite{van2017neural} and is trained with a reconstruction loss. To enhance smoothness, we apply Lipschitz regularization~\cite{liu2022learning} to the encoder. Fig.~\ref{fig:lipfq-vae architecture} provides an overview of the proposed action tokenizer.

In a VAE~\cite{zhao2023learning}, an encoder \( q(\mathbf{z}|\mathbf{x}) \) maps the input action $\mathbf{x}$ to a latent vector $\mathbf{z}\in\mathcal{Z}$. Here, $\mathbf{x}$ corresponds to the action represented by the relative position and angle of gripper poses, following the common practice in RoboCasa~\cite{nasiriany2024robocasa} and ManiSkill~\cite{tao2024maniskill3}. A decoder \( p(\mathbf{x}|\mathbf{z}) \) then reconstructs action $\hat{\mathbf{x}}$ from the latent space. The VAE is trained by optimizing a reconstruction loss and a KL divergence regularization term to enforce a well-structured latent space~\cite{van2017neural}. Nevertheless, VAE's continuous latent space poses challenges for autoregressive models, which are more effective in dealing with discrete variables, such as vocabulary~\cite{zhang2022generative}.

To enable discrete representation, we employ VQ-VAE~\cite{van2017neural}, which only differs by encoding the latent space via vector quantization. Specifically, the posterior distribution is defined over a codebook \( \mathcal{Z} = \{\mathbf{e}_1, \mathbf{e}_2, \dots, \mathbf{e}_K\} \subset \mathbb{R}^{K \times D} \), where \( K \) is the number of entries in the codebook and $D$ is the latent dimension. Each latent vector is mapped to the nearest codebook entry using the L2 distance:
\begin{equation}
\mathbf{z} = \mathbf{e}_k, \quad k = \arg\min_j \|q_{\theta}(\mathbf{x}) - \mathbf{e}_j\|_2^2,
\end{equation}
where \( q_\theta(\mathbf{x}) \) is a neural network-based approximation of the posterior distribution, \textit{i.e.}, the encoder output.



Though VQ-VAE offers effective compression, its latent space lacks smoothness, making temporal preservation challenging~\cite{zhao2023learning}. To address this, we propose an improved VQ-VAE variant that enhances temporal smoothness in action representation. Inspired by Liu \textit{et al.}~\cite{liu2022learning}, we constrain the Lipschitz condition on the encoder to ensure the smoothness of the latent space. A neural network \( q_\theta \) is $c$-Lipschitz continuous if:
\begin{equation}
\label{eq: lipschitz-condition}
    \| q_{\theta}(\mathbf{t}_0) - q_{\theta}(\mathbf{t}_1) \|_\lambda \leq c \|\mathbf{t}_0 - \mathbf{t}_1 \|_\lambda,
\end{equation}
for all inputs \( \mathbf{t}_0, \mathbf{t}_1 \) under a chosen \( \lambda \)-norm, where \( c \) is the Lipschitz bound. 

Theoretical results in~\cite{miyato2018spectral} show that, for a fully connected network with 1-Lipschitz activation functions (e.g., ReLU), the Lipschitz bound \( c \) can be estimated as $c = \prod \| \mathbf{W}^{(\ell)} \|_\lambda,$ where $\mathbf{W}^{(\ell)}$ is the weight of the network at ${\ell}$-th layer. Hence, the network's Lipschitz bound is the product of the per-layer bounds, making \textit{weight normalization} an effective method to enforce the Lipschitz condition~\cite{song2023lipsnet}. Selecting an appropriate \( \lambda \)-norm for measuring the Lipschitz bound in Eq.~\eqref{eq: lipschitz-condition} can significantly influence the normalization scheme~\cite{miyato2018spectral}. We adopt the approach of~\cite{liu2022learning} and employ the \( \infty \)-norm, which normalizes each row \( \mathbf{W}^{(\ell)}_i \) such that its absolute row sum does not exceed a predefined bound.  To this end, each layer $\ell$ in the encoder \( q_\theta \) is augmented with a trainable Lipschitz bound \( c_\ell \), introduced for every row \( i \) as follows:

\begin{equation}
    \label{eq: regularization}
    \mathbf{W}_i^{(\ell)} = \frac{\mathbf{W}_{i}^{(\ell)}}{\sum_j |{\mathbf{W}_{i,j}^{(\ell)}}|} \cdot \text{softplus}(c_\ell),
\end{equation}  
where \( \text{softplus}(c_\ell) = \ln(1 + e^{c_\ell}) \) enforces positivity of the Lipschitz bounds via reparameterization. In practice, since \( c_\ell \) is typically large, we have \( c_\ell \approx \text{softplus}(c_\ell) \). If the absolute row sum is already smaller than \( \text{softplus}(c_\ell) \), no rescaling is applied. This formulation guarantees that the Lipschitz bound per layer remains bounded by \( \text{softplus}(c_\ell) \), thereby preventing excessive weight clipping during training. As illustrated in Fig.~\ref{fig:lipfq-vae architecture}, we apply the regularization in Eq.~\eqref{eq: regularization} after each linear layer and before each ReLU activation.

\nbf{Training Loss.} We train LipVQ-VAE jointly with the ICRT framework (Sec.~\ref{subsec: ICRT}). We follow the training procedure in~\cite{van2017neural} to train a well-structured and stable latent space. The training objective comprises three key loss terms:
\begin{itemize}[leftmargin=*]
    \item  The reconstruction loss $\mathcal{L}_{\text{reconstruction}} = \|\hat{\mathbf{x}} - \mathbf{x}\|_2^2$ ensures a meaningful latent space with accurate reconstruction.
    \item The codebook loss $\mathcal{L}_{\text{codebook}} = \|  \text{sg}[q_\theta(\mathbf{x})]-\mathbf{z} \|_2^2,$  where \(\text{sg}[\cdot]\) denotes the stop-gradient operation, updates the codebook by moving towards the encoder outputs. 
    \item The commitment loss $\mathcal{L}_{\text{commit}} = \|   q_\theta(\mathbf{x}) -\text{sg}[\mathbf{z}]\|_2^2,$  prevents the codebook underfit problem~\cite{esser2021taming}.
\end{itemize}
Finally, similar to~\cite{liu2022learning}, we include a regularization term for the Lipschitz bound per layer in Eq.~\eqref{eq: regularization} as $\mathcal{L}_{\text{Lipschitz}} = \prod\text{softplus}(c_\ell)$. The overall objective function is:
\begin{equation}
    \mathcal{L} = \mathcal{L}_{\text{reconstruction}} + \alpha\mathcal{L}_{\text{codebook}} + \beta\mathcal{L}_{\text{commit}} + \gamma\mathcal{L}_{\text{Lipschitz}},
\end{equation}
where the weighting coefficients $\alpha$, $\beta$, and $\gamma$ are set to $1.0$, $0.25$, and $10^{-6}$, respectively, following~\cite{van2017neural, liu2022learning}.

\section{EXPERIMENT}
\label{sec: experiment}

We conduct experiments in both simulation and the real world to assess the effectiveness of the LipVQ-VAE action tokenizer. 
Our experiments aim to answer the following questions:

\begin{itemize}[leftmargin=*]
    \item \textbf{Q1:} How much can LipVQ-VAE benefit ICIL?
    \item \textbf{Q2:} Does Lipschitz regularization improve the smoothness of action representation, and how does it affect the ICIL?
    \item \textbf{Q3:} Can our method transfer from simulation to real world?
\end{itemize}

\subsection{Simulation Robotic Manipulation Results.}
\nbf{Experimental Setup.} 
We perform simulation experiments in RoboCasa~\cite{nasiriany2024robocasa} and ManiSkill~\cite{tao2024maniskill3}, and all experiments are conducted using an NVIDIA RTX 4090 GPU.
For RoboCasa, we assess $7$ foundational skill families and train for 500K iterations, following the standard setting~\cite{nasiriany2024robocasa}. For ManiSkill, we evaluate $3$ tasks: \textit{Pick Cube}, \textit{Push Cube}, and \textit{Stack Cube}. We train all methods for 30K iterations, aligning with the baseline~\cite{tao2024maniskill3}.
Policy performance is measured using success rate, following the practice in each simulator~\cite{nasiriany2024robocasa, tao2024maniskill3}.

To assess the overall ICRT framework, we compare our method against the following existent baselines: \textit{i)} BC-Transformer~\cite{mandlekar2021matters}, \textit{ii)} ACT~\cite{zhao2023learning}, and \textit{iii)} MCR~\cite{jiang2025robots}. 
In the original paper~\cite{zhao2023learning}, ACT employs multiple encoder-decoder transformer layers and large feed-forward dimension, exceeding our GPU memory; thus, we downscale ACT's model size to match BC-Transformer for fair comparison.

Within the ICRT framework (Sec.~\ref{subsec: ICRT}), we compare the proposed LipVQ-VAE against five action tokenizers, 
including: \textit{i)} MLP~\cite{fu2024context}, \textit{ii)} discrete bin action\cite{brohan2022rt}, \textit{iii)} FAST~\cite{pertsch2025fast}, \textit{iv)} VQ-VAE~\cite{van2017neural}, and \textit{v)} LFQ-VAE~\cite{yu2024language}. 
For the FAST tokenizer, we fine-tune 1M actions from RoboCasa and ManiSkill, aligning with the DROID dataset~\cite{khazatsky2024droid}.

\begin{figure*}[!ht]
    \centering
    \includegraphics[width=1.0\linewidth]{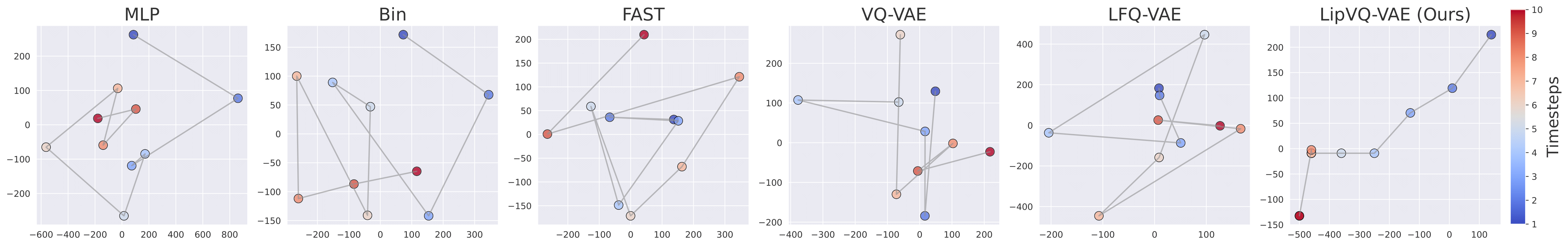}
    \caption{\textbf{Latent trajectories visualization.}  Using t-SNE, we visualize latent representations of different action tokenizers given the same action trajectory.}
    \label{fig:qualitative-results}
\end{figure*}

\nbf{Results on RoboCasa.} 
We evaluate all baselines on the MimicGen~\cite{mandlekar2023mimicgen} dataset. 
Table~\ref{table: robocasa} reports the results, revealing the following key insights: \textit{i)} In-context learning methods improve the performance. 
Specifically, integrating our LipVQ-VAE action tokenizer into ICRT achieves a 5.3\% improvement over BC-Transformer, the state-of-the-art method on RoboCasa to our knowledge. 
\textit{ii)} Action tokenization plays a crucial role in ICRT’s effectiveness, as evidenced by an 8.8\% performance gap between the best (LipVQ-VAE) and worst (MLP) tokenizers. Notably, an ineffective tokenizer can degrade performance, with MLP underperforming BC-Transformer by 3.5\%. 
\textit{iii)} Among all action tokenizers, LipVQ-VAE achieves the highest performance, surpassing the second-best tokenizer, LFQ-VAE, by 4.1\%, demonstrating the effectiveness of Lipschitz regularization in preserving smoothness and enhancing performance.

\begin{table}[!ht]
    \centering
    \caption{Robotic manipulation results on RoboCasa.}\label{table: robocasa}
    \vspace{1ex}
    \setlength{\tabcolsep}{1pt} 
    \resizebox{\linewidth}{!}{
        \begin{tabular}{@{}rcccccccc@{}}
            \toprule
            \diagbox{Baseline}{Task} & \shortstack{Pick \\ and \\ Place} & \shortstack{Open \\ Close \\ Doors} & \shortstack{Open \\ Close \\ Drawers} & \shortstack{Turning \\ Levers \\ $ $} & \shortstack{Twisting \\ Knobs \\ $ $} & \shortstack{Insertion \\ $ $ \\ $ $} & \shortstack{Pressing \\ Buttons \\ $ $} & \shortstack{\textbf{Avg.} \\ $ $ \\ $ $}\\
            \midrule
            MCR~\cite{jiang2025robots}  &  0.00 & 0.31 & 0.18 & 0.17 & 0.02 & 0.01 & 0.22 & 0.120  \\
            ACT~\cite{zhao2023learning} & 0.01 & 0.13 & 0.17 & 0.15 & 0.12 & 0.07 & 0.06 & 0.083 \\
            
            BC-Transformer~\cite{mandlekar2021matters} & 0.29 & 0.55 & 0.78 & 0.62 & 0.31 & 0.24 & \textbf{0.78} & 0.477\\
            \midrule
            ICRT~\cite{fu2024context}+MLP~\cite{fu2024context} & 0.20 & 0.61 & 0.81 & 0.70 & 0.32 & 0.35 & 0.64 & 0.442 \\
            ICRT+Bin~\cite{brohan2022rt} & 0.25 & 0.75 & 0.78 & 0.81 & 0.32 & 0.34 & 0.59 & 0.483\\
            ICRT+FAST~\cite{pertsch2025fast}  & 0.30 & 0.59 & 0.80 & 0.57 & 0.39 & 0.19 & 0.63 & 0.471\\
            ICRT+VQ-VAE~\cite{van2017neural} & 0.20 & 0.70 & 0.84 & 0.77 & 0.27 & 0.18 & 0.70 & 0.475\\
            ICRT+LFQ-VAE~\cite{yu2024language} & 0.27 & 0.69 & 0.83 & \textbf{0.77} & 0.40 & 0.27 & 0.68 & 0.489\\
            \midrule
            ICRT+LipVQ-VAE (ours) & \textbf{0.32} & \textbf{0.80} & \textbf{0.84} & 0.68 & \textbf{0.41} & \textbf{0.41} & 0.59 & \textbf{0.530 }\\
            \bottomrule
        \end{tabular}
    }
\end{table}

\nbf{Results on ManiSkill.} 
To evaluate the generalization of action tokenizers beyond ICIL, we replace ACT's cVAE action encoder with different action tokenizers.
Since MCR does not pretrains on RGB-D~\cite{jiang2025robots}, we extend a depth channel to the image backbone and train jointly with the policy head. 
Table~\ref{table: maniskill} presents the results, which align with our findings on RoboCasa. 
LipVQ-VAE still achieves the highest performance among all action tokenizers, exceeding LFQ-VAE by 6\% 
While both ACT and our LipVQ-VAE address the action smoothing problem, the original ACT method does not explicitly regularize the latent features, resulting in a 19.7\% performance gap in favor of our method.

\begin{table}[!ht]
    \centering
    \caption{Results on ManiSkill.}\label{table: maniskill}
    \vspace{1ex}
    \setlength{\tabcolsep}{5pt} 
    \resizebox{\linewidth}{!}{
        \begin{tabular}{rcccc}
            \toprule
            \diagbox{Baseline}{Task} & Pick cube & Push cube & Stack Cube & \textbf{Avg.}\\
            \midrule
            MCR~\cite{jiang2025robots} & 0.56 & 0.51 & 0.11 & 0.393 \\
            ACT~\cite{zhao2023learning} & 0.20 & 0.76 & 0.30 & 0.420 \\
            BC-Transformer~\cite{mandlekar2021matters} & 0.04 & \textbf{0.98} & 0.14 & 0.387 \\
            \midrule
            ACT+Bin~\cite{brohan2022rt} & 0.71 & 0.52 & 0.25 & 0.493 \\
            ACT+FAST~\cite{pertsch2025fast} & 0.70 & 0.48 & 0.25 & 0.477 \\
            ACT+VQ-VAE~\cite{van2017neural} & 0.64 & 0.80 & 0.21 & 0.550 \\
            ACT+LFQ-VAE~\cite{yu2024language} & 0.74 & 0.70 &  0.23 & 0.557 \\
            \midrule
            ACT+LipVQ-VAE (ours) & \textbf{0.78} & 0.77 & \textbf{0.30} & \textbf{0.617}\\
            \bottomrule
        \end{tabular}
    }
\end{table}

\nbf{Smoothness Evaluation.} To evaluate the smoothness of the proposed method, we extract 500 action latent features from the RoboCasa experiment using different action tokenizers and compute the least energy as the smoothness score based on the theory of~\cite{horn1983curve}.
This metric quantifies energy as the integral of the square of the curvature traversing the trajectory. The less energy required to traverse between timesteps, the smoother the representation. 
Table~\ref{table:least_energy} reports the results, revealing a clear positive correlation—smoother action tokenization leads to higher manipulation success rates, in connection with Table~\ref{table: robocasa}. Among all tokenizers, LipVQ-VAE achieves the best smoothness score (0.63). While the bin action tokenizer exhibits non-smooth behavior, it performs well on RoboCasa, likely due to its use of integer values, which amplifies differences between action latent vectors.

\begin{table}[!ht]
    \centering
    \caption{Smoothness Quantitative Results.}
    \label{table:least_energy}
    \vspace{1ex}
    \setlength{\tabcolsep}{3pt} 
    \resizebox{\linewidth}{!}{
        \begin{tabular}{lcccccc}
            \toprule
            Tokenizer & \shortstack{ MLP~\cite{fu2024context}} & \shortstack{Bin\\ ~\cite{brohan2022rt}} & \shortstack{FAST\\ ~\cite{pertsch2025fast}} & \shortstack{VQ-VAE \\ ~\cite{van2017neural}} & \shortstack{LFQ-VAE\\ ~\cite{yu2024language}}  & \shortstack{LipVQ-VAE \\ (Ours)} \\
            \midrule
            Smoothness $\downarrow$ & 5.71 & 12.57 & 5.67 & 4.79 & 2.34 & \textbf{0.63} \\
            \bottomrule
        \end{tabular}
    }
\end{table}

Additionally, we visualize the latent vectors tokenized by each method in Fig.~\ref{fig:qualitative-results}, applying t-SNE to project the first 10 timesteps of an action trajectory into 2D space. The qualitative results confirm that LipVQ-VAE produces smoother latent representations than other tokenizers. 
These findings underscore the effectiveness of Lipschitz regularization and highlight smoothness as a key factor in improving action tokenization for ICIL.

\subsection{Ablation Study}
This section evaluates key components of the proposed framework. We first examine the impact of Lipschitz regularization on the action tokenizer, followed by an analysis of codebook size in the LipVQ-VAE latent space. Finally, we assess the in-context learning framework's transferability to unseen domains.


\nbf{Impact of Lipschitz Regularization.} We analyze the effect of adding Lipschitz regularization to action tokenizers in the RoboCasa simulator as in Table~\ref{table: lipschitz-regularization-impact}. Among the five action tokenizers, LFQ-VAE, Bin, and FAST use a discrete latent space (integers or binary vectors), making Lipschitz regularization infeasible due to non-continuity. Our results show that incorporating Lipschitz regularization consistently improves performance by 2.3\% for the MLP tokenizer and 5.5\% for VQ-VAE. Furthermore, VQ-VAE outperforms MLP in both regularized and non-regularized settings, suggesting that quantization into discrete vectors enables a more structured representation of the observation-action space.


\begin{table}[!ht]
    \centering
    \begin{minipage}{0.545\linewidth}
        \centering
        \caption{Lipschitz regularization's impact}
        \label{table: lipschitz-regularization-impact}
        \setlength{\tabcolsep}{1pt} 
        \vspace{3pt}
        \resizebox{\linewidth}{!}{
        \begin{tabular}{l|c|c}
            \toprule
            Baseline & \shortstack{Lipschitz \\ Reg.} & \shortstack{Success \\ Rate} \\
            \midrule
            ICRT~\cite{fu2024context}+MLP~\cite{fu2024context} & \ding{55} & 0.442 \\
            ICRT~\cite{fu2024context}+MLP~\cite{fu2024context} & \ding{51} & \underline{0.465} \\
            \midrule
            ICRT~\cite{fu2024context}+VQ-VAE~\cite{van2017neural} & \ding{55} & 0.475 \\
            ICRT~\cite{fu2024context}+VQ-VAE~\cite{van2017neural} & \ding{51} (ours) & \textbf{0.530} \\
            \bottomrule
        \end{tabular}
        }
    \end{minipage}
    \hfill
    \begin{minipage}{0.444\linewidth}
        \centering
        \caption{Sim-to-real success rates}
        \label{tab:sim2real}
        \setlength{\tabcolsep}{1pt} 
        \vspace{3pt}
        \resizebox{\linewidth}{!}{
        \begin{tabular}{lcc}
            \toprule
            \diagbox{Baseline}{Task} &  \shortstack{Open \\ Close \\ Drawers} &  \shortstack{Open \\ Close \\ Doors} \\
            \midrule
            BC-Transformer \cite{mandlekar2021matters}        & 0.02 & 0.02 \\ \midrule
            ICRT~\cite{fu2024context}+Bin \cite{brohan2022rt}             & 0.04 & 0.02 \\ \midrule
            ICRT+LipVQ-VAE (ours)& \textbf{0.12} & \textbf{0.14} \\
            \bottomrule
        \end{tabular}
        }
    \end{minipage}
\end{table}

\begin{figure}[!ht]
    \centering
    \includegraphics[width=1.0\linewidth]{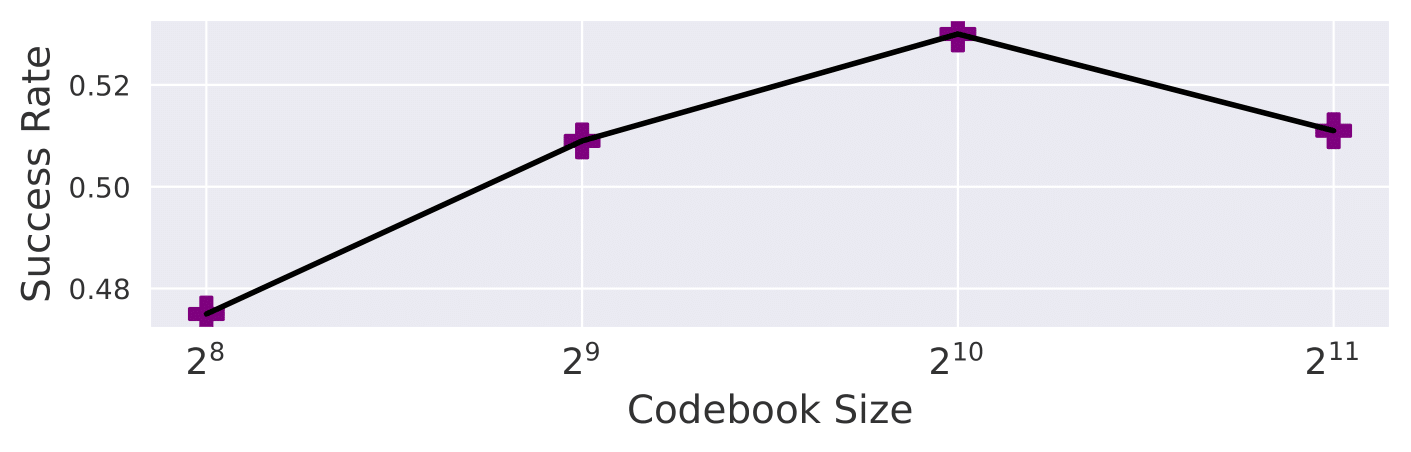}
    \caption{\textbf{Impact of LipVQ-VAE's codebook size in RoboCasa.}}
    \label{fig:impact-codebook}
\end{figure}

\nbf{Impact of Codebook Size.} We vary the VQ-VAE codebook size from $2^8$ to $2^{11}$ as in Fig.~\ref{fig:impact-codebook} and observe improved performance in RoboCasa with a larger codebook size. The codebook size of $2^{10}$ yields the best performance (0.530), as further increases introduce additional latency without yielding performance gains.

\nbf{Cross-dataset Results.} To evaluate the generalization capabilities in a new dataset, we evaluate the performance transferring from the MimicGen~\cite{mandlekar2023mimicgen} dataset to the human dataset presented in~\cite{nasiriany2024robocasa}. The human dataset could feature sparser object arrangements since MimicGen generates diverse high-level kitchen environments using LLMs~\cite{nasiriany2024robocasa}. Table~\ref{table: cross-dataset} reports the results. Notably, BC-Transformer underperforms compared to other in-context learning baselines, suggesting that the ICRT framework better adapts to new contexts. Even with the weakest action tokenizer (MLP), ICRT outperforms BC-Transformer by 4.9\%. In addition, the LipVQ-VAE tokenizer enjoys better performance within the ICRT framework, improving other action tokenizers by 2.3\%.

\begin{table}[!ht]
    \centering
    \caption{Cross-dataset Robotic Manipulation.}\label{table: cross-dataset}
    \vspace{1ex}
    \setlength{\tabcolsep}{1pt} 
    \resizebox{\linewidth}{!}{
        \begin{tabular}{rcccccccc}
            \toprule
            & \multicolumn{8}{c}{\textbf{MimicGen}~\cite{mandlekar2023mimicgen}$\to$\textbf{Human}~\cite{nasiriany2024robocasa}} \\ \midrule
            \diagbox{Baseline}{Task} & \shortstack{Pick \\ and \\ Place} & \shortstack{Open \\ Close \\  Doors} & \shortstack{Open \\ Close \\ Drawers} & \shortstack{Turning \\ Levers \\ $ $} & \shortstack{Twisting \\ Knobs \\ $ $} & \shortstack{Insertion \\ $ $ \\ $ $} & \shortstack{Pressing \\ Buttons \\ $ $}  & \shortstack{\textbf{Avg.} \\ $ $ \\ $ $} \\
            \midrule
            MCR~\cite{jiang2025robots} & 0.00 & 0.24 & 0.20 & 0.12 & 0.00 & 0.00 & 0.19 & 0.096 \\
            ACT~\cite{zhao2023learning} & 0.00 & 0.10 & 0.15 & 0.13 & 0.09 & 0.08 & 0.05 & 0.066 \\
            BC-Transformer~\cite{mandlekar2021matters} & 0.25 & 0.41 & 0.73 & 0.62 & 0.28 & 0.18 & 0.64 & 0.408 \\
            \midrule
            ICRT~\cite{fu2024context}+MLP~\cite{fu2024context}~\cite{fu2024context} & 0.21 & 0.61 & \textbf{0.87} & 0.77 & 0.30 & 0.36 & 0.52 & 0.457 \\
            ICRT+Bin~\cite{brohan2022rt} & 0.26 & 0.75 & 0.79 & 0.74 & 0.31 & 0.29 & 0.60 & 0.495 \\
            ICRT+FAST~\cite{pertsch2025fast} & 0.30  & 0.63 & 0.77 & 0.74 & 0.36 & 0.39 & 0.42 & 0.481 \\
            ICRT+VQ-VAE~\cite{van2017neural} & 0.21 & 0.70 & 0.83 & 0.77 & 0.36 & 0.31 & 0.60 & 0.483 \\
            ICRT+LFQ-VAE~\cite{yu2024language} & 0.29  & 0.65 & 0.82 & \textbf{0.79} & 0.38 & 0.28 & 0.62 & 0.503 \\
            \midrule
            ICRT+LipVQ-VAE (ours) & \textbf{0.32} & \textbf{0.76} & 0.83 & 0.71 & \textbf{0.38} & \textbf{0.42} & \textbf{0.64} & \textbf{0.525} \\
            \bottomrule
        \end{tabular}
    }
\end{table}

\nbf{Zero-shot Manipulation Results.} To further evaluate generalization in new tasks and objects, we conduct a zero-shot experiment by training on eight demo atomic tasks and evaluating the remaining tasks, following the split by~\cite{nasiriany2024robocasa}. Table~\ref{table: zero-shot-robocasa} presents the results, indicating that \textit{i)} In-context learning significantly enhances zero-shot performance compared to non-in-context baselines, with our approach surpassing BC-Transformer by 2.4\%, and \textit{ii)} LipVQ-VAE outperforms other tokenizers by at least 0.2\%, highlighting the benefits of temporal smoothness for unseen action trajectories.

\begin{table}[!ht]
    \centering
    \caption{Zero-shot Robotic Manipulation Results.}\label{table: zero-shot-robocasa}
    \vspace{1ex}
    \setlength{\tabcolsep}{1pt} 
    \resizebox{\linewidth}{!}{
        \begin{tabular}{rcccccccc}
            \toprule
            \diagbox{Baseline}{Task} & \shortstack{Pick \\ and \\ Place} & \shortstack{Open \\ Close \\  Doors} & \shortstack{Open \\ Close \\ Drawers} & \shortstack{Turning \\ Levers \\ $ $} & \shortstack{Twisting \\ Knobs \\ $ $} & \shortstack{Insertion \\ $ $ \\ $ $} & \shortstack{Pressing \\ Buttons \\ $ $}  & \shortstack{\textbf{Avg.} \\ $ $ \\ $ $} \\
            \midrule
            MCR~\cite{jiang2025robots} & 0.00 & 0.00 & 0.00  & 0.06 & 0.04 & 0.01 & 0.00 & 0.021  \\
            ACT~\cite{jiang2025robots} & 0.00 & 0.00 & 0.00  & 0.05 & 0.04 & 0.01 & 0.00 & 0.019  \\
            BC-Transformer~\cite{mandlekar2021matters} & 0.00 & 0.00 & 0.00 & 0.10 & 0.20 & 0.00 & 0.00 & 0.028 \\
            \midrule
            ICRT~\cite{fu2024context}+MLP~\cite{fu2024context} & 0.00 & 0.01 & 0.00 & 0.19 & 0.22 & 0.01 & 0.02 & 0.047 \\
            ICRT+Bin~\cite{brohan2022rt} & 0.01 & 0.01 & 0.00 & 0.16 & 0.24 & 0.01 & 0.02 & 0.046 \\
            ICRT+FAST~\cite{pertsch2025fast} & 0.00 & 0.00 & 0.00 & 0.20 & 0.12 & 0.01 & 0.00 & 0.041 \\
            ICRT+VQ-VAE~\cite{van2017neural} & 0.00 & 0.00 & 0.00 & 0.16 & 0.20 & 0.01 & 0.02 & 0.041 \\
            ICRT+LFQ-VAE~\cite{yu2024language} & 0.01 & \textbf{0.01} & 0.00 & 0.24 & 0.14 & \textbf{0.02} & 0.02 & 0.049 \\
            \midrule
            ICRT+LipVQ-VAE (ours) & \textbf{0.01} & 0.00 & 0.00 & \textbf{0.25} & \textbf{0.24} & 0.00 & \textbf{0.02} & \textbf{0.052} \\
            \bottomrule
        \end{tabular}
    }
\end{table}

\subsection{Real-world Robotic Manipulation Results.}
We collect demonstrations in RoboCasa using a Kinova Gen3 arm for two tasks: \emph{open/close} a drawer and \emph{open/close} a door. Each task has 10 human-teleoperated demonstrations gathered via a 3Dconnexion SpaceMouse. To expand coverage, we employ MimicGen to generate an additional 2000 synthetic demonstrations, yielding a large-scale dataset with Kinova Gen3 in simulation. We record two camera viewpoints (an external third-person view and an eye-in-hand camera), and we synchronize the simulation controller and the real Kinova controller at \SI{50}{Hz}. We then collect a separate set of 10 human demonstrations in a real environment to evaluate sim-to-real performance. These real demonstrations (also augmented via MimicGen), combined with our simulation-based data, form a unified dataset used to train several imitation-learning methods. 

We compare our proposed ICRT+LipVQ-VAE against BC-Transformer and a variant of ICRT using bin-based discretization. We train each method and evaluate a total of 50 demonstrations per task. Table~\ref{tab:sim2real} summarizes the sim-to-real success rates on the drawer-opening and drawer-closing tasks. Despite the large-scale augmented dataset, direct sim-to-real transfer remains challenging. Our ICRT+LipVQ-VAE approach achieves smoother motion and notably higher success, reaching around 12\% on real hardware, while the baseline approaches struggle to exceed 2\%. A video of the experiments can be seen in the supplementary material. 

\begin{figure}[ht]
\centering
\scalebox{0.9}{
\def\svgwidth{1\columnwidth}
\begingroup%
  \makeatletter%
  \providecommand\color[2][]{%
    \errmessage{(Inkscape) Color is used for the text in Inkscape, but the package 'color.sty' is not loaded}%
    \renewcommand\color[2][]{}%
  }%
  \providecommand\transparent[1]{%
    \errmessage{(Inkscape) Transparency is used (non-zero) for the text in Inkscape, but the package 'transparent.sty' is not loaded}%
    \renewcommand\transparent[1]{}%
  }%
  \providecommand\rotatebox[2]{#2}%
  \newcommand*\fsize{\dimexpr\f@size pt\relax}%
  \newcommand*\lineheight[1]{\fontsize{\fsize}{#1\fsize}\selectfont}%
  \ifx\svgwidth\undefined%
    \setlength{\unitlength}{745.38395439bp}%
    \ifx\svgscale\undefined%
      \relax%
    \else%
      \setlength{\unitlength}{\unitlength * \real{\svgscale}}%
    \fi%
  \else%
    \setlength{\unitlength}{\svgwidth}%
  \fi%
  \global\let\svgwidth\undefined%
  \global\let\svgscale\undefined%
  \makeatother%
  \begin{picture}(1,0.59144623)%
    \lineheight{1}%
    \setlength\tabcolsep{0pt}%
    \put(0,0){\includegraphics[width=\unitlength,page=1]{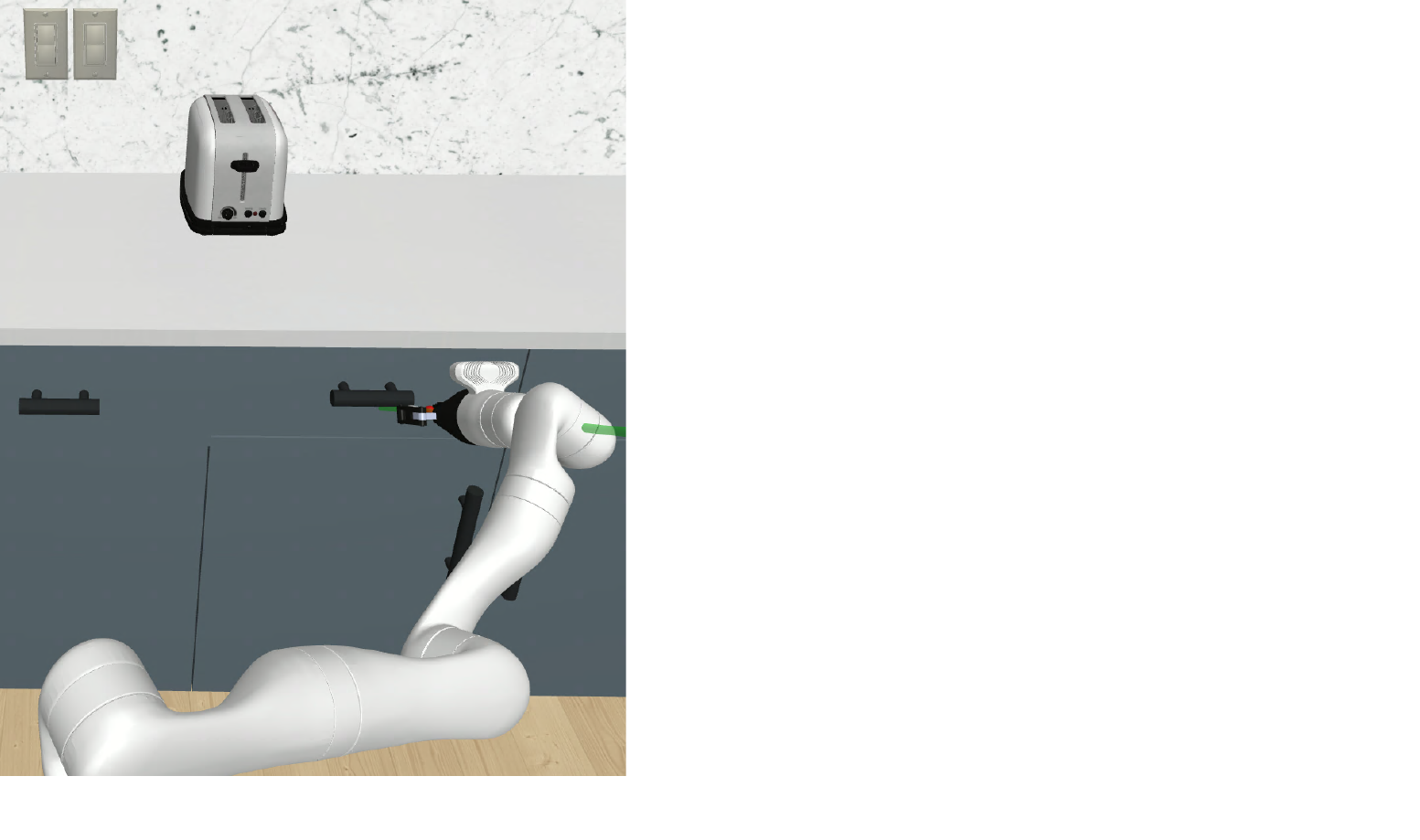}}%
    \put(0.19033456,0.00530607){\color[rgb]{0,0,0}\makebox(0,0)[lt]{\lineheight{1.25}\smash{\begin{tabular}[t]{l}(a)\end{tabular}}}}%
    \put(0.7130132,0.00717115){\color[rgb]{0,0,0}\makebox(0,0)[lt]{\lineheight{1.25}\smash{\begin{tabular}[t]{l}(b)\end{tabular}}}}%
    \put(0,0){\includegraphics[width=\unitlength,page=2]{experiments.pdf}}%
  \end{picture}%
\endgroup%

}
\vspace{1ex}
\caption{(a) Data generation in RoboCasa. (b) Real-world experiment}
\label{fig:robot_experiment}
\end{figure}

\subsection{Discussion}
Experiments on RoboCasa (Table~\ref{table: robocasa}) and ManiSkill (Table~\ref{table: maniskill}) demonstrate that LipVQ-VAE significantly outperforms other baselines (\textbf{Q1}). Smoothness evaluation (Table~\ref{table:least_energy}) and ablation study (Table~\ref{table: lipschitz-regularization-impact}) reveal a strong correlation between smoothness and success rate, indicating that Lipschitz regularization effectively enhances action smoothness and improves performance (\textbf{Q2}). Finally, real robot experiments (Table~\ref{tab:sim2real}) confirm that our method transfers successfully to real-world applications (\textbf{Q3}).

Our approach still sees some limitations. First, similar to other action tokenizers, LipVQ-VAE does not explicitly tackle the challenge of integrating multiple modalities, including vision, language, proprioception, and action. Effectively fusing action representation into a shared latent space with multi-modal observations remains an open problem that this work does not fully resolve. Consequently, our approach's zero-shot performance remains modest (Table~\ref{table: zero-shot-robocasa}) despite its goal to enhance generalization. Finally, while ICIL performs well in simulation, its real-world performance is limited, highlighting persistent sim-to-real challenges~\cite{nasiriany2024robocasa}.

\section{CONCLUSION}
This paper introduces LipVQ-VAE, an action tokenizer that enforces the Lipschitz condition in the latent quantized space to enhance smoothness in action representation. We show that LipVQ-VAE improves smoothness over other tokenizers, leading to improved robotic manipulation performance due to greater stability in the latent space. Integrated into an in-context imitation learning framework, LipVQ-VAE outperforms state-of-the-art baselines on RoboCasa and ManiSkill. Real-world experiments confirm its successful transfer from simulation to physical robots.

\bibliographystyle{class/IEEEtran}
\bibliography{class/IEEEabrv,reference}

\begin{thebibliography}{10}
\providecommand{\url}[1]{#1}
\csname url@rmstyle\endcsname
\providecommand{\newblock}{\relax}
\providecommand{\bibinfo}[2]{#2}
\providecommand\BIBentrySTDinterwordspacing{\spaceskip=0pt\relax}
\providecommand\BIBentryALTinterwordstretchfactor{4}
\providecommand\BIBentryALTinterwordspacing{\spaceskip=\fontdimen2\font plus
\BIBentryALTinterwordstretchfactor\fontdimen3\font minus \fontdimen4\font\relax}
\providecommand\BIBforeignlanguage[2]{{%
\expandafter\ifx\csname l@#1\endcsname\relax
\typeout{** WARNING: IEEEtran.bst: No hyphenation pattern has been}%
\typeout{** loaded for the language `#1'. Using the pattern for}%
\typeout{** the default language instead.}%
\else
\language=\csname l@#1\endcsname
\fi
#2}}

\bibitem{mandlekar2021matters}
A.~Mandlekar, D.~Xu, J.~Wong, S.~Nasiriany, C.~Wang, R.~Kulkarni, L.~Fei-Fei, S.~Savarese, Y.~Zhu, and R.~Mart{\'\i}n-Mart{\'\i}n, ``What matters in learning from offline human demonstrations for robot manipulation,'' in \emph{CoRL}, 2021.

\bibitem{team2024octo}
O.~M. Team, D.~Ghosh, H.~Walke, K.~Pertsch, K.~Black, O.~Mees, S.~Dasari, J.~Hejna, T.~Kreiman, C.~Xu, \emph{et~al.}, ``Octo: An open-source generalist robot policy,'' \emph{arXiv preprint arXiv:2405.12213}, 2024.

\bibitem{o2024open}
O’Neill \emph{et~al.}, ``Open x-embodiment: Robotic learning datasets and rt-x models: Open x-embodiment collaboration 0,'' in \emph{ICRA}, 2024.

\bibitem{khazatsky2024droid}
A.~Khazatsky \emph{et~al.}, ``{DROID:} {A} large-scale in-the-wild robot manipulation dataset,'' in \emph{RSS}, 2024.

\bibitem{di2024keypoint}
N.~Di~Palo and E.~Johns, ``Keypoint action tokens enable in-context imitation learning in robotics,'' \emph{arXiv preprint arXiv:2403.19578}, 2024.

\bibitem{xia2024kinematic}
W.~Xia, D.~Wang, X.~Pang, Z.~Wang, B.~Zhao, D.~Hu, and X.~Li, ``Kinematic-aware prompting for generalizable articulated object manipulation with llms,'' in \emph{ICRA}, 2024.

\bibitem{vosylius2025instant}
V.~Vosylius and E.~Johns, ``Instant policy: In-context imitation learning via graph diffusion,'' in \emph{ICLR}, 2025.

\bibitem{wang2024large}
X.~Wang, W.~Zhu, M.~Saxon, M.~Steyvers, and W.~Y. Wang, ``Large language models are latent variable models: Explaining and finding good demonstrations for in-context learning,'' \emph{NeurIPS}, 2024.

\bibitem{fu2024context}
L.~Fu, H.~Huang, G.~Datta, L.~Y. Chen, W.~C.-H. Panitch, F.~Liu, H.~Li, and K.~Goldberg, ``In-context imitation learning via next-token prediction,'' \emph{arXiv preprint arXiv:2408.15980}, 2024.

\bibitem{park2025incontext}
C.~F. Park, A.~Lee, E.~S. Lubana, Y.~Yang, M.~Okawa, K.~Nishi, M.~Wattenberg, and H.~Tanaka, ``In-context learning of representations,'' in \emph{ICLR}, 2025.

\bibitem{wang2025skil}
S.~Wang, J.~You, Y.~Hu, J.~Li, and Y.~Gao, ``Skil: Semantic keypoint imitation learning for generalizable data-efficient manipulation,'' \emph{arXiv preprint arXiv:2501.14400}, 2025.

\bibitem{pertsch2025fast}
K.~Pertsch, K.~Stachowicz, B.~Ichter, D.~Driess, S.~Nair, Q.~Vuong, O.~Mees, C.~Finn, and S.~Levine, ``Fast: Efficient action tokenization for vision-language-action models,'' \emph{arXiv preprint arXiv:2501.09747}, 2025.

\bibitem{zheng2024prise}
R.~Zheng, C.-A. Cheng, H.~D. III, F.~Huang, and A.~Kolobov, ``{PRISE}: {LLM}-style sequence compression for learning temporal action abstractions in control,'' in \emph{ICML}, 2024.

\bibitem{zhao2023learning}
T.~Zhao, V.~Kumar, S.~Levine, and C.~Finn, ``Learning fine-grained bimanual manipulation with low-cost hardware,'' \emph{RSS}, 2023.

\bibitem{song2023lipsnet}
X.~Song, J.~Duan, W.~Wang, S.~E. Li, C.~Chen, B.~Cheng, B.~Zhang, J.~Wei, and X.~S. Wang, ``Lipsnet: a smooth and robust neural network with adaptive lipschitz constant for high accuracy optimal control,'' in \emph{ICLR}, 2023.

\bibitem{brohan2022rt}
A.~Brohan, N.~Brown, J.~Carbajal, Y.~Chebotar, J.~Dabis, C.~Finn, K.~Gopalakrishnan, K.~Hausman, A.~Herzog, J.~Hsu, \emph{et~al.}, ``Rt-1: Robotics transformer for real-world control at scale,'' \emph{arXiv preprint arXiv:2212.06817}, 2022.

\bibitem{wang2024smooth}
W.~Wang, J.~Duan, X.~Song, L.~Xiao, L.~Chen, Y.~Wang, B.~Cheng, and S.~E. Li, ``Smooth filtering neural network for reinforcement learning,'' \emph{T-IV}, 2024.

\bibitem{yang2024vq}
C.~Yang, D.~Liconti, and R.~K. Katzschmann, ``Vq-ace: Efficient policy search for dexterous robotic manipulation via action chunking embedding,'' \emph{arXiv preprint arXiv:2411.03556}, 2024.

\bibitem{fujimoto2024sale}
S.~Fujimoto, W.-D. Chang, E.~Smith, S.~S. Gu, D.~Precup, and D.~Meger, ``For sale: State-action representation learning for deep reinforcement learning,'' \emph{NeurIPS}, 2024.

\bibitem{vaswani2017attention}
A.~Vaswani, ``Attention is all you need,'' \emph{NeurIPS}, 2017.

\bibitem{yu2024language}
L.~Yu, J.~Lezama, N.~B. Gundavarapu, L.~Versari, K.~Sohn, D.~Minnen, Y.~Cheng, A.~Gupta, X.~Gu, A.~G. Hauptmann, \emph{et~al.}, ``Language model beats diffusion-tokenizer is key to visual generation,'' in \emph{ICLR}, 2024.

\bibitem{bharadhwaj2024roboagent}
H.~Bharadhwaj, J.~Vakil, M.~Sharma, A.~Gupta, S.~Tulsiani, and V.~Kumar, ``Roboagent: Generalization and efficiency in robot manipulation via semantic augmentations and action chunking,'' in \emph{ICRA}, 2024.

\bibitem{mysore2021regularizing}
S.~Mysore, B.~Mabsout, R.~Mancuso, and K.~Saenko, ``Regularizing action policies for smooth control with reinforcement learning,'' in \emph{ICRA}, 2021.

\bibitem{liu2022learning}
H.-T.~D. Liu, F.~Williams, A.~Jacobson, S.~Fidler, and O.~Litany, ``Learning smooth neural functions via lipschitz regularization,'' in \emph{SIGGRAPH}, 2022.

\bibitem{van2017neural}
A.~Van Den~Oord, O.~Vinyals, \emph{et~al.}, ``Neural discrete representation learning,'' \emph{NeurIPS}, 2017.

\bibitem{ma2024survey}
Y.~Ma, Z.~Song, Y.~Zhuang, J.~Hao, and I.~King, ``A survey on vision-language-action models for embodied ai,'' \emph{arXiv preprint arXiv:2405.14093}, 2024.

\bibitem{duan2022temporal}
C.~Duan, J.~Ding, S.~Chen, Z.~Yu, and T.~Huang, ``Temporal effective batch normalization in spiking neural networks,'' \emph{NeurIPS}, 2022.

\bibitem{nasiriany2024robocasa}
S.~Nasiriany, A.~Maddukuri, L.~Zhang, A.~Parikh, A.~Lo, A.~Joshi, A.~Mandlekar, and Y.~Zhu, ``Robocasa: Large-scale simulation of household tasks for generalist robots,'' in \emph{RSS}, 2024.

\bibitem{tao2024maniskill3}
S.~Tao, F.~Xiang, A.~Shukla, Y.~Qin, X.~Hinrichsen, X.~Yuan, C.~Bao, X.~Lin, Y.~Liu, T.-k. Chan, \emph{et~al.}, ``Maniskill3: Gpu parallelized robotics simulation and rendering for generalizable embodied ai,'' \emph{arXiv preprint arXiv:2410.00425}, 2024.

\bibitem{brown2020language}
T.~Brown, B.~Mann, N.~Ryder, M.~Subbiah, J.~D. Kaplan, P.~Dhariwal, A.~Neelakantan, P.~Shyam, G.~Sastry, A.~Askell, \emph{et~al.}, ``Language models are few-shot learners,'' \emph{NeurIPS}, 2020.

\bibitem{mirchandani2023large}
S.~Mirchandani, F.~Xia, P.~Florence, B.~Ichter, D.~Driess, M.~G. Arenas, K.~Rao, D.~Sadigh, and A.~Zeng, ``Large language models as general pattern machines,'' in \emph{Conference on Robot Learning}.\hskip 1em plus 0.5em minus 0.4em\relax PMLR, 2023, pp. 2498--2518.

\bibitem{kwon2024language}
T.~Kwon, N.~Di~Palo, and E.~Johns, ``Language models as zero-shot trajectory generators,'' \emph{IEEE Robotics and Automation Letters}, 2024.

\bibitem{vosylius2023few}
V.~Vosylius and E.~Johns, ``Few-shot in-context imitation learning via implicit graph alignment,'' in \emph{CoRL}, 2023.

\bibitem{caron2021emerging}
M.~Caron, H.~Touvron, I.~Misra, H.~J{\'e}gou, J.~Mairal, P.~Bojanowski, and A.~Joulin, ``Emerging properties in self-supervised vision transformers,'' in \emph{Proceedings of the IEEE/CVF international conference on computer vision}, 2021, pp. 9650--9660.

\bibitem{xu2024flow}
M.~Xu, Z.~Xu, Y.~Xu, C.~Chi, G.~Wetzstein, M.~Veloso, and S.~Song, ``Flow as the cross-domain manipulation interface,'' in \emph{CoRL}, 2024.

\bibitem{zhang2025dynamics}
X.~Zhang, S.~Liu, P.~Huang, W.~J. Han, Y.~Lyu, M.~Xu, and D.~Zhao, ``Dynamics as prompts: In-context learning for sim-to-real system identifications,'' \emph{RA-L}, 2025.

\bibitem{chandak2019learning}
Y.~Chandak, G.~Theocharous, J.~Kostas, S.~Jordan, and P.~Thomas, ``Learning action representations for reinforcement learning,'' in \emph{ICML}, 2019.

\bibitem{zech2019action}
P.~Zech, E.~Renaudo, S.~Haller, X.~Zhang, and J.~Piater, ``Action representations in robotics: A taxonomy and systematic classification,'' \emph{IJRR}, 2019.

\bibitem{shafiullah2022behavior}
N.~M. Shafiullah, Z.~Cui, A.~A. Altanzaya, and L.~Pinto, ``Behavior transformers: Cloning $ k $ modes with one stone,'' \emph{NeurIPS}, 2022.

\bibitem{kim2024openvla}
M.~J. Kim, K.~Pertsch, S.~Karamcheti, T.~Xiao, A.~Balakrishna, S.~Nair, R.~Rafailov, E.~Foster, G.~Lam, P.~Sanketi, \emph{et~al.}, ``Openvla: An open-source vision-language-action model,'' \emph{arXiv preprint arXiv:2406.09246}, 2024.

\bibitem{huang2024emotion}
P.~Huang, Y.~Hu, N.~Nechyporenko, D.~Kim, W.~Talbott, and J.~Zhang, ``Emotion: Expressive motion sequence generation for humanoid robots with in-context learning,'' \emph{arXiv preprint arXiv:2410.23234}, 2024.

\bibitem{alwani2022decore}
M.~Alwani, Y.~Wang, and V.~Madhavan, ``Decore: Deep compression with reinforcement learning,'' in \emph{CVPR}, 2022.

\bibitem{allshire2021laser}
A.~Allshire, R.~Mart{\'\i}n-Mart{\'\i}n, C.~Lin, S.~Manuel, S.~Savarese, and A.~Garg, ``Laser: Learning a latent action space for efficient reinforcement learning,'' in \emph{ICRA}, 2021.

\bibitem{watson2023inferring}
J.~Watson and J.~Peters, ``Inferring smooth control: Monte carlo posterior policy iteration with gaussian processes,'' in \emph{CoRL}, 2023.

\bibitem{styrud2024bebop}
J.~Styrud, M.~Mayr, E.~Hellsten, V.~Krueger, and C.~Smith, ``Bebop-combining reactive planning and bayesian optimization to solve robotic manipulation tasks,'' in \emph{ICRA}, 2024.

\bibitem{he2016deep}
K.~He, X.~Zhang, S.~Ren, and J.~Sun, ``Deep residual learning for image recognition,'' in \emph{CVPR}, 2016.

\bibitem{radford2021learning}
A.~Radford, J.~W. Kim, C.~Hallacy, A.~Ramesh, G.~Goh, S.~Agarwal, G.~Sastry, A.~Askell, P.~Mishkin, J.~Clark, \emph{et~al.}, ``Learning transferable visual models from natural language supervision,'' in \emph{ICML}, 2021.

\bibitem{zhang2022generative}
D.~Zhang, N.~Malkin, Z.~Liu, A.~Volokhova, A.~Courville, and Y.~Bengio, ``Generative flow networks for discrete probabilistic modeling,'' in \emph{ICML}, 2022.

\bibitem{miyato2018spectral}
T.~Miyato, T.~Kataoka, M.~Koyama, and Y.~Yoshida, ``Spectral normalization for generative adversarial networks,'' in \emph{ICML}, 2018.

\bibitem{esser2021taming}
P.~Esser, R.~Rombach, and B.~Ommer, ``Taming transformers for high-resolution image synthesis,'' in \emph{CVPR}, 2021.

\bibitem{jiang2025robots}
G.~Jiang, Y.~Sun, T.~Huang, H.~Li, Y.~Liang, and H.~Xu, ``Robots pre-train robots: Manipulation-centric robotic representation from large-scale robot dataset,'' in \emph{ICLR}, 2025.

\bibitem{mandlekar2023mimicgen}
A.~Mandlekar, S.~Nasiriany, B.~Wen, I.~Akinola, Y.~Narang, L.~Fan, Y.~Zhu, and D.~Fox, ``Mimicgen: A data generation system for scalable robot learning using human demonstrations,'' in \emph{CoRL}, 2023.

\bibitem{horn1983curve}
B.~K. Horn, ``The curve of least energy,'' \emph{TOMS}, 1983.

\end{thebibliography}

\end{document}